\documentclass[10pt,twocolumn,letterpaper]{article}

\usepackage[final]{cvpr}
\usepackage{times}
\usepackage{epsfig}
\usepackage{graphicx}
\usepackage{amsmath}
\usepackage{amssymb}
\usepackage{makecell}
\usepackage{tablefootnote}
\usepackage{multirow}
\usepackage{hyperref}
\usepackage{booktabs}
\usepackage{soul}
\usepackage{subcaption}
\usepackage{algorithm}
\usepackage{bm}
\usepackage{algpseudocode}

\algrenewcommand\algorithmicrequire{\textbf{Input:}}
\algrenewcommand\algorithmicensure{\textbf{Output:}}
\begin{document}

%%%%%%%%% TITLE
\title{Balacing Beyond Discrete Categories: Continuous Demographic Labels for Fair Face Recognition}

\author{\textbf{Pedro C. Neto}\\
INESC TEC \& FEUP\\
{\tt\small pedro.carneiro.neto.21@gmail.com}
\and
\textbf{Naser Damer}\\
Fraunhofer Institute for Computer Graphics Research IGD\\
Technische Universitat Darmstadt\\
\and
\textbf{Jaime S. Cardoso}\\
FEUP \& INESC TEC\\
\and
\textbf{Ana F. Sequeira}\\
INESC TEC \\
}

\maketitle
\thispagestyle{empty}

%%%%%%%%% ABSTRACT
\begin{abstract}
   Bias has been a constant in face recognition models. Over the years, researchers have looked at it from both the model and the data point of view. However, their approach to mitigation of data bias was limited and lacked insight on the real nature of the problem. Here, in this document, we propose to revise our use of ethnicity labels as a continuous variable instead of a discrete value per identity. We validate our formulation both experimentally and theoretically, showcasing that not all identities from one ethnicity contribute equally to the balance of the dataset; thus, having the same number of identities per ethnicity does not represent a balanced dataset. We further show that models trained on datasets balanced in the continuous space consistently outperform models trained on data balanced in the discrete space. We trained more than 65 different models, and created more than 20 subsets of the original datasets. 
\end{abstract}

%%%%%%%%% BODY TEXT
\section{Introduction}
\label{sec:intro}

Face recognition (FR) has grown over the recent years. Deep learning (DL)-based FR systems are currently available for diverse purposes, ranging from border control to mobile authentication devices. While the performance of the FR systems has been advancing over time, it was deep learning that enabled outstanding improvements in the realm of recognition with face biometrics. These improvements and the apparent nearly flawless performance led to the implementation and deployment of face recognition systems in critical domains. Many of these deployments have led to negative consequences due to wrong predictions, lack of explainability and, most importantly, notable biases regarding certain demographic groups. This raised concerns on both researchers and users, and it is a problem that is still occurring~\cite{bacchini2019race,krishnapriya2020issues,Neto2023CompressedModels}. Hence, these flawed systems keep on been negatively mentioned by news outlets\footnote{\url{https://www.scientificamerican.com/article/police-facial-recognition-technology-cant-tell-black-people-apart/}}, often being associated with wrong predictions in court-related tasks\footnote{\url{https://capitalbnews.org/facial-recognition-wrongful-arrests/}}~\cite{johnson2022facial}.

\begin{figure}
    \centering
    \includegraphics[width=\linewidth]{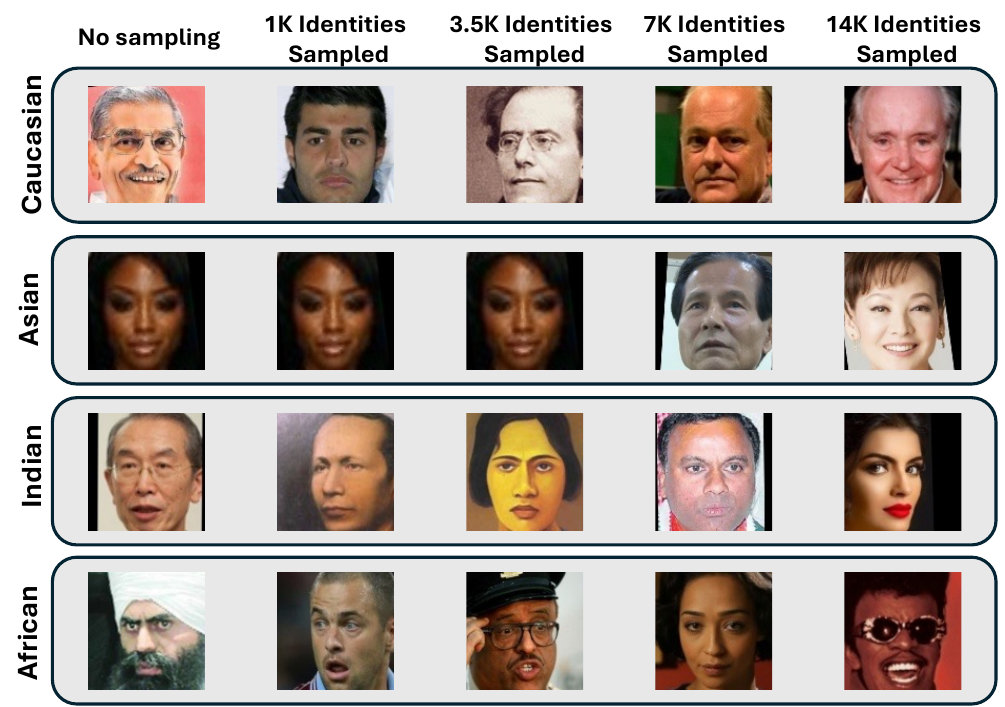}
    \caption{\textbf{Demonstration of the Sampling Effect:} For each sampling intensity (using sampling strategy A) the sample with the lowest confidence of each label is shown. For instance, on No Sampling, it is possible to see an Asian image (according to the dataset label) that the ethnicity classifier model has a low confidence in the Asian class. It is also noticeable that as the sampling intensifies, the ethnicity traits become more prominent.  }
    \label{fig:lest_representative}
\end{figure}

Deep learning success has been driven by three important elements. First, the improvements in computational power propelled the ability to run computationally expensive methods that were previously disregarded due to their computational cost~\cite{Krizhevsky2012AlexNet}. Additionally, new techniques allowed for an increased depth in neural networks' architectures~\cite{he2016deep}. Thus, the complexity of DL models has grown significantly. Finally, the ability to scrap large scale datasets from the web increased the quantity of data available to train these deep systems~\cite{Guo2016Ms1M,ZhuWebFace260M}. The first two elements came at the cost of transparency~\cite{NetoCausality_Inspited}, while the non-rigorous collection of data led to unbalanced datasets that fed the demographic biases found in face recognition models. Together, they create an environment that favours the presence of undetected biases.

The relation of a biased dataset and a biased model is well-studied, and evidence indicates this relationship also can be found in face recognition tasks. In 2021, Wang~\textit{et al.}~\cite{wang2021meta} has proposed two face recognition datasets. The first of these datasets follows an ethnicity distribution similar to the real-world distribution, whereas the other balanced the number of identities per each ethnicity. The same authors have also proposed RFW, a balanced dataset for benchmark and bias detection~\cite{Wang2019RFW}. Similarly, Wu and Bowyer~\cite{Wu2023BADataset} investigated which elements should be balanced on a balanced dataset and proposed a test dataset, sampled from VGGFace2~\cite{Cao2018VGGFace2}, that is balanced for head-pose, quality and brightness. BFW~\cite{Robinson2020BFW} dataset was also sampled from VGGFace2, and balanced its data on both gender and ethnicity. In general, these datasets focused on balancing the discrete labels that identify each gender/ethnicity or any other attribute. However, in practice, one can pose the question "How Asian is my Asian sample?", as the prevalence of certain ethnicity traits is not equal in all individuals. Hence, we argue that not all training samples from a certain demographic trait contribute equally to the discriminative power of a model on test samples from that trait.

\begin{table*}[ht]
    \centering
    \footnotesize
    \caption{Methods for Dataset Balancing in Bias Mitigation}
    \begin{tabular}{|l|l|l|}
        \hline
        \textbf{Method} & \textbf{Approach} & \textbf{Type of Ethnicity labels} \\
        \hline
        Wang \textit{et al.}~\cite{Wang2019RFW} & Introduced Racial Faces in-the-wild (RFW) & Discrete \\
        \hline
        Robinson \textit{et al.}~\cite{Robinson2020BFW} & Introduced Balanced Faces in-the-wild (BFW) & Discrete \\
        \hline
        Gwilliam \textit{et al.}~\cite{Gwilliam2021RethinkingAssumptions} & Challenged dataset balance assumptions (\#IDs per Ethnicity)& Discrete \\
        \hline
        Yucer \textit{et al.}~\cite{Yucer2020CycleGanAugment} & Ethnicity augmentation with CycleGAN & Discrete \\
        \hline
        Chandaliya \textit{et al.}~\cite{chandaliya2024towards} & Ethnicity alteration for a balanced dataset &  Discrete\\
        \hline
        Wang \textit{et al.}~\cite{Wang2020EffectiveSampling} & Oversampling strategy for rare classes (Not on Biometrics) & Discrete \\
        \hline
    \end{tabular}
    \label{tab:bias_mitigation_methods}
\end{table*}

Having proper representativeness in the dataset is essential to achieve a sufficiently good discriminative function. Thus, based on the previously described balancing strategy, we defined a set of novel strategies that provide grounds for a stronger selection of identities with respect to each ethnicity. These strategies aim to share, across research fields, that discrete labels-based balancing might yield a sub-optimal result with respect to the fairness of the model across demographic groups. With this in mind, we propose a set of steps to balance data with respect to a demographic attribute while using continuous labels. For this, we start by proposing a method based on a state-of-the-art FR system to create these continuous labels. Additionally, we leverage these labels to assess the true balance of the dataset, showcasing the unbalanced nature of the data. Finally, we propose three different strategies to leverage these labels to create a new set of samples with a higher balance. We validate our strategies by removing 3.5\%, 12.5\%, 25\% and 50\% of the identities in the training dataset. Figure~\ref{fig:lest_representative} shows the least representative identity of each group at a given point of one of the sampling strategies.  We compare the performance of our strategy versus models trained on the original dataset and also on a set constructed by randomly removing the same number of identities that retain the balance of the discrete labels. Our results highlight the ability of the three approaches in the reduction of two different bias metrics. Additionally, in terms of accuracy, models trained on some identity-reduced sets created with our approach surpass or are on par with the models trained on the entire dataset. These results were acquired for multiple backbones and losses.

We can summarise our contributions as follows: 

\begin{itemize}
    \item We question the sanity of discrete labels in demographic balancing through the use of continuous labels, we analyse discretely balanced datasets, showcasing their unbalance on the continuous space;
    \item We propose three sampling strategies to balance datasets in the continuous space;
    \item We release all the ``datasets'' created with these strategies and recommend their use;
    \item  We showcase, through experiments with different backbones and losses, that models trained on these datasets balanced in the continuos space improve fairness metrics and accuracy when compared to models trained on similarly sized datasets with discrete balancing. 
\end{itemize}

Beside this introductory section and the conclusion, we organise this paper into four main sections. We start by discussing the current literature on the topics of face recognition and fairness within this context. The following section presents the main contributions with a discussion regarding the ethnicity classifier, the notation, and the formulation of the sampling strategies. Before presenting and discussing the results in the last section, we introduce the experimental setup with all the necessary details to reproduce this work. The provided datasets and some of the trained models can be found at \footnote{\url{https://github.com/NetoPedro/Continuous-Ethnicity-Face-Recognition}} and \footnote{\url{https://huggingface.co/collections/netopedro/continuous-ethnicity-face-recognition-683d775e507954149965e5b6}}.

\section{Related work}
\label{sec:related}

\textbf{Face Recognition:} Face Recognition tasks are normally designated as Verification and Identification. The former handles 1-1 comparisons where a reference image is given and the objective is to classify a probe image as belonging to the same person, or not. The latter task requires 1-N comparisons to match the input face to an identity within a given gallery. In order to be able to conduct these matching processes, the models utilised require a high discriminative power to mitigate both false positives and false negatives. With the rise of deep learning, contrastive losses, such as Triplet Loss, gained momentum and achieved state-of-the-art results~\cite{Schroff2015FaceNet,Cao2018VGGFace2}. These works learn to distinguish different faces by ensuring that the distance between an anchor and a same identity is smaller than the distance of the anchor to a sample from a different identity.  Recently, Softmax-based losses have gained traction as they do not suffer from the network collapse problems associated with contrastive methods~\cite{Levi2021TripletCollapse}. In 2018, Wang~\textit{et al.}~\cite{wang2018cosface} proposed CosFace, where they proposed an additive margin to the resulting cosine of the angle between weights and features. Furthermore, CosFace normalises the weights and features to have norm 1, further scaling it to a constant $s$. Deng~\textit{et al.}~\cite{deng2022ArcFace} extended CosFace by adding the margin to the angle between features and weights, and not to the resulting cosine. Boutros~\textit{et al.}~\cite{boutros2022ElasticFace} studied variations of both ArcFace and CosFace, where the margin is modelled by a random variable sampled, for each sample-epoch combination, from a Gaussian distribution. Huang~\textit{et al.}~\cite{huang2020Curricular}, inspired by the principles of curriculum learning, proposed CurricularFace, a novel loss that tackles samples from the easiest to the hardest as training progresses.

\textbf{Bias in Face Recognition:} The differences in the performance of Face Recognition with respect to certain demographic groups have been discussed in commercial systems by Boulamwini and Gebru~\cite{Buolamwini2018GenderShades}, highlighting superior performance on males with lighter skin tones. Additionally, this work proposed one of the first balanced evaluation datasets. The repercussions of such findings led to updates on commercial systems (e.g. IBM and Microsoft)~\cite{Raji2019ActionableAuditing}. Over time, the problem of biases in commercial systems has been widely discussed~\cite{drozdowski2020demographic,Dooley2022RobustnessDisparities,Jaiswal2022TwoFace}. Huber~\textit{et al.}~\cite{huber2023explainability} noted that explanation maps reflect the hidden biases of a model that they aim to explain. 

Over the years, researchers have approached the mitigation of this problem from two different perspectives. First, mitigating the bias that results from the dataset~\cite{Gwilliam2021RethinkingAssumptions}. Secondly, understanding and reducing the bias resulting from the deep architecture chosen~\cite{Dooley2023RethinkingBias}. Despite claims of favouring one research direction, research efforts have been allocated to both directions. 

Gong~\textit{et al.}~\cite{Gong2021AdaptativeBias} proposed a learnable adaptive set of masks for convolutional filters. Different demographic images are split into different groups, and for each group, a set of masks is learnt in order to enhance the prevalence of ethnicity-specific features in the final embedding. Similar to the previous work, Wang~\textit{et al.}~\cite{wang2021meta} method leveraged the differences of the different demographic groups to learn independent margins for each of these groups. Huang~\textit{et al.}~\cite{Huang2023GradientAttention} explores the gradient activation maps of the different ethnicity groups. Their proposed method leverages adversarial learning to create race-invariant gradient activation maps. Although their reported bias decreases, it has lower overall performance in all four ethnicities when compared to~\cite{wang2021meta}. Serna~\textit{et al.}~\cite{serna2022sensitive} modified the triplet generation process of the triplet loss. Their restricted semi-hard selection selects triplets from the same demographic group in order to increase the discriminative capabilities intra-group. 

After hypothesising that demographic bias could originate from the architecture, Dooley~\textit{et al.}~\cite{Dooley2023RethinkingBias} developed a Neural Architecture Search pipeline to find an architecture and a set of hyperparameters that minimise the bias of the final model. Neto~\textit{et al.}~\cite{Neto2023CompressedModels} investigated the behaviour of the model's bias before and after quantisation on several architectures and datasets. Their findings have shown that quantisation with synthetic data has the potential to mitigate bias to a certain degree. 

SensitiveNets~\cite{Morales2021SensitiveNets} propose a bias mitigation strategy that arises from an increase in the privacy and protection of sensitive attributes. Hence, their approach maps the feature vector to a space where these attributes are suppressed. Leveraging a different perspective, Gong~\textit{et al.}~\cite{Gong2020JointlyDebiasing} interprets demographic information as an important piece of the debiasing process. Hence, their work proposes a novel architecture that does disentangled prediction of identity, gender, race and age. To ensure no mutual information between the different predictions, an adversarial strategy was adopted. This latter strategy has a negative impact on the overall performance of the model. 

To validate their proposed unsupervised approach to bias mitigation, Wang~\textit{et al.}~\cite{Wang2019RFW} introduced the validation dataset Racial Faces in-the-wild (RFW). This dataset has been used over the past six years to evaluate both commercial and research systems. Two years later, Robinson~\textit{et al.}~\cite{Robinson2020BFW} introduced Balanced Faces in-the-wild (BFW), a benchmark dataset that balances for ethnicity and gender. Despite balancing for more demographics, this dataset has roughly 4x fewer faces and 15x fewer identities.

Gwilliam~\textit{et al.}~\cite{Gwilliam2021RethinkingAssumptions} challenged prior assumptions that a balanced dataset leads to higher bias mitigation. We believe that the previous definition is outdated, and we further show that models trained on datasets balanced with continuous labels (which are imbalanced if we consider discrete labels), perform better than models trained on datasets balanced for discrete labels. Hence, we also reformulate the assumption of Gwilliam~\textit{et al.}~\cite{Gwilliam2021RethinkingAssumptions} as our current datasets are not truly balanced.

DeAndres{-}Tame~\textit{et al.}~\cite{Tame2024FRCSyn} hosted a competition aimed at evaluating face recognition system trained on synthetic data and evaluated on the different ethnicity groups. Yucer~\textit{et al.}~\cite{Yucer2020CycleGanAugment} introduced a solution that consisted of two steps. First, resampling from VGGFace2~\cite{Cao2018VGGFace2} a set of 1200 identities balanced for race. Afterwards, their approach would generate versions of the same identity with its ethnicity flipped. Improvements came at the cost of overall performance. Similar experiments have been conducted by Chandaliya~\textit{et al.}~\cite{chandaliya2024towards} to construct a more challenging and balanced dataset.

Wang~\textit{et al.}~\cite{Wang2020EffectiveSampling} studied a sampling strategy in a study regarding different bias mitigation strategies. Their sampling approach focuses on increasing the likelihood that the model sees a sample from a rare class through oversampling. Our proposed approach differs significantly as we propose to downsample  the data to approach a more balanced configuration.

\section{Methodology}
\label{sec:meth}

In this paper, we propose a novel theoretical framework to interpret ethnicity labels as values belonging to a continuous spectrum. This further defies current notions of data balancing and equilibrium with respect to ethnicity. As such, we show that not all samples from a demographic group are equally strong in their representation of its group. Hence, we theorize that equal number of samples/identities per group does not suffice nor contributes exclusively to a balanced dataset. Rather, we believe that the level of belonging to a demographic group of each sample directly influences the balancing act. To validate our approach we propose three sampling approaches to balance face recognition datasets for ethnicity labels in the continuous space. These sampling approaches, which would not be possible in a discrete formulation, leverage the flexibility of the novel interpretation of ethnicity labels. 

\begin{figure}[h!]
    \centering
    \includegraphics[width=1.01\linewidth]{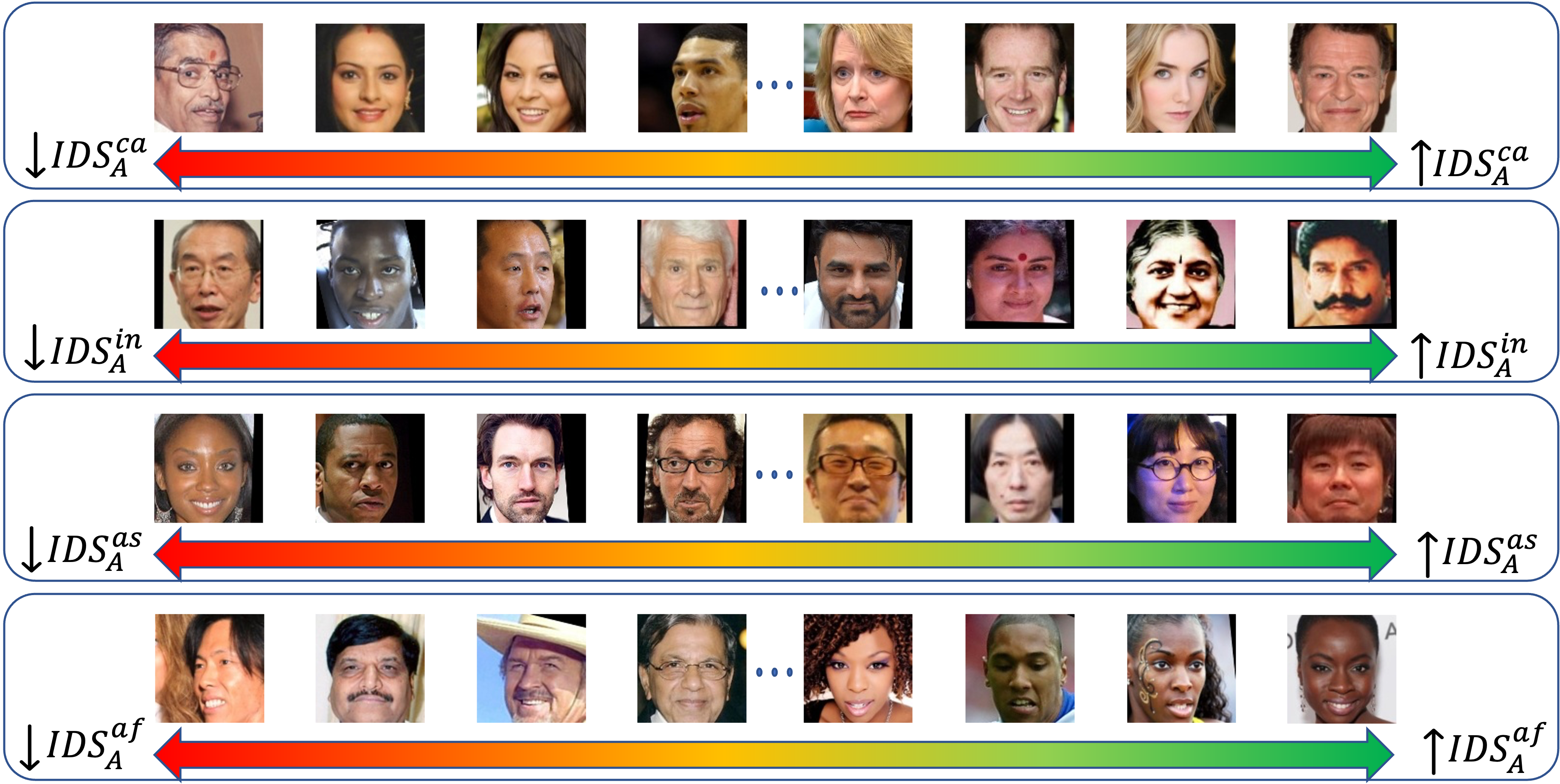}
    \caption{\textbf{Ethnicity Shift in Demographic Groups:} given the labels of the dataset, we show one sample of the identities that are the least and the most aligned with their label (introduced later in Section~\ref{sec:notation}). It is visible the presence of noise in this label, as well as the high variability even within samples with high scores and the non-descriptive nature of discrete labels. }
    \label{fig:image_strenghts}
\end{figure}

 Figure~\ref{fig:image_strenghts} highlights, in each row, the inter-ethnicity variability for each of the four distinct demographic groups. On each row, from left to the right one can observe the images with lowest to the highest prevalence of ethnicity specific traits. As shown by the four left images of each row, one can verify not only the existence of possibly noisy labels, but also non-representative samples of each ethnicity and samples that could belong to more than one of the four demographic groups, a clear representation of the continuous nature of demographic labels.

In this section we discuss the methodology used to verify the previously mentioned hypothesis. Given a function that maps from the image space to a four-dimensional continuous space, which indicates the degree of belonging of an image to four different ethnicities. We show, individually for each ethnicity, that removing strongly-scored samples degrades the performance much faster than removing weakly-scored ones. Then, considering the label distribution in the continuos space, we show that by evening these distributions one can achieve similar performance with unbalanced number of identities per ethnicity group, and far less identities on each group. In fact, we show that this balancing improves fairness metrics even if we do not consider the number of images per identity, nor the number of identities per ethnicity group, which further increases the unbalance of discrete labels.

\subsection{Balancing Training Data}
Given a face recognition dataset where each identity is labelled with an ethnicity from the set \( \mathcal{D} \), where \( |\mathcal{D}| = d \), and an ethnicity score agent \( f: \mathbb{R}^{h \times w} \to \mathbb{R}^d \) that infers the degree of belonging of an image to each of the \( d \) ethnicity labels in \( \mathcal{D} \), one can produce image-specific continuos ethnicity labels. In this direction, identity-specific scores can be computed from these image score and a final representativeness score can be given to each ethnicity from the identity scores. To assess the overall representativeness of an ethnicity and the identity score, we designed a set of possible methods with different focuses and trade-offs. The first of these scores, can be viewed as comparative proxy between ethnicities to infer the balancing of the dataset, whereas the identity score highlights the contributions of each identity to their ethnicity score.

To validate our theory, we conducted several experiments. Following different sampling strategies, we show that having the same number of identities per ethnicity group does not mean that the dataset is balanced.

\subsubsection{Notation}
\label{sec:notation}
Before presenting our approach to balance a face recognition dataset, it is necessary to introduce some information regarding the variables and the notation to be used. For the remaining of this paper, let us consider $d=4$. As such, we present the ethnicity label $y$, where $y \in \mathcal{D}$ and $\mathcal{D} =\{Af, As, In, Ca\}$. In addition, we introduce three variables that are associated with the ethnicity score at three different granularity levels, $ies$ for the image score, $IDS$ for the identity score and $ES$ for the ethnicity score. The image score, $ies_{j,i} \in \mathbb{R} ^4 $, can be defined as follows: 

\begin{align}
    ies_{j,i} = \begin{pmatrix}
    \vspace{0.2cm}
    ies^{Af}_{j,i}\\
    \vspace{0.2cm}
    ies^{Ca}_{j,i}\\
    \vspace{0.2cm}
    ies^{As}_{j,i}\\
    \vspace{0.2cm}
    ies^{In}_{j,i}
    \end{pmatrix}, 
\end{align}

where $j$ and $i$, both scalars, represent the identity of the image and the image itself, respectively. Within the vector, it is possible to observe four entries which correspond to the image ethnicity score with respect to each ethnicity group. The identity score $IDS_{j} \in \mathbb{R} ^4 $ is defined as:

\begin{align}
    IDS_{j} = \begin{pmatrix}
    \vspace{0.2cm}
    IDS^{Af}_{j}\\
    \vspace{0.2cm}
    IDS^{Ca}_{j}\\
    \vspace{0.2cm}
    IDS^{As}_{j}\\
    \vspace{0.2cm}
    IDS^{In}_{j}
    \end{pmatrix} 
\end{align}

In a similar fashion, each entry of the vector corresponds to the identity ethnicity score with respect to each ethnicity. Finally, $ES \in \mathbb{R} ^{4 \times 4} $ is a matrix defined as:

\begin{align}
    ES_{} = \begin{pmatrix}
    \vspace{0.2cm}
    ES^{Af}_{Af} & ES^{Ca}_{Af} & ES^{As}_{Af} & ES^{In}_{Af} \\
    \vspace{0.2cm}
    ES^{Af}_{Ca} & ES^{Ca}_{Ca} & ES^{As}_{Ca} & ES^{In}_{Ca}\\
    \vspace{0.2cm}
    ES^{Af}_{As} & ES^{Ca}_{As} & ES^{As}_{As} & ES^{In}_{As}\\
    \vspace{0.2cm}
    ES^{Af}_{In} & ES^{Ca}_{In} & ES^{As}_{In} & ES^{In}_{In}
    \end{pmatrix}
\end{align}

$ES^{Af}_{Ca}$ refers to the ethnicity score for all the identities labelled as “Caucasians” with respect to how representative they are of the “African” group. In the context of this research, we focus our attention on the ethnicity score calculated with respect to the ground truth ethnicity. Hence, $ES$ can be simplified as $diag(ES)$, where we refer to each element as $diag(ES)^{(y)}$.

%NAser done till here
\subsubsection{Protocol A}

Two elements of strong importance for the performance of face recognition systems are the number of identities and of images per identity in its training data. We hypothesized that our balancing approach should work regardless of our focus on the number of identities, and images per identity. As such, we have designed a sampling protocol that completely ignores both these elements, only focusing on the ethnicity score. This behaviour implies that an identity with 100 images could be removed in favour of an identity with ten images, if the second had a higher score. On this Protocol, A, for an image $i$ from identity $j$ with an image ethnicity score vector $ies_{j,i}$, the identity ethnicity score vector, $IDS_{Aj}$, can be calculated as 

\vspace{-0.15cm}
\begin{align}
     IDS_{Aj} = \sum^{M_j}_i \frac{ies_{j,i}}{M_j},
\end{align}

where $M_j$ represents the number of images from the identity $j$. From this score, we get the average ethnicity score of this identity. Its formulation, as an average, ignores the impact of the number of images associated with an identity. On this protocol, we can further calculate the Ethnicity Score $ES_{Ay}$, for a demographic group $y$ with $N_y$ identities as
\vspace{-0.15cm}
\begin{align}
     ES_{Ay} = \sum^{N_y}_j \frac{IDS_{Aj}}{N_y}. 
\end{align}

This score ignores the number of identities of a given ethnicity group due to its averaging over all identities. Given this ethnicity score, the ethnicity imbalance is quite significant, as shown in subsequent sections. $IDS^{(y)}_{A}$ can be used to rank all identities within a demographic group $y$. Following this, one can remove the identity $j$ with the lowest $IDS^{(y)}_{Aj}$ from the ethnicity $y$ with the smaller value for $diag(ES_A)^{(y)}$. This will lead to an increase in the latter score. This is shown in Algorithm~\ref{alg:protocolA}.

\begin{algorithm}
\caption{Sampling Process for Protocol A}
\label{alg:protocolA}
\begin{algorithmic}
\Require $Z \geq 0$
\State $\mathcal{E} \gets \{Af, As, Ca, In\}$ \Comment{Ethnic Groups: African, Asian, Caucasian, Indian}
\State $N \gets 0$
\While{$N \neq Z$}
    \For{$y \in \mathcal{E}$}
        \State $ES_A(y) \gets 0$  \Comment{Initialize ethnicity-level score}
        \State $K_y \gets 0$ \Comment{Ethnicity count for sample $j$}
        \For{$j \in y$} 
            \State $IDS_A(j) \gets 0$ \Comment{Initialize ID-level score}
            \State $M_j \gets 0$ \Comment{Image count for sample $j$}
            \For{$img \in j$}
                \State $IDS_A(j) \gets IDS_A(j) + ies(img)$
                \State $M_j \gets M_j + 1$
            \EndFor
            \State $IDS_A(j) \gets \frac{IDS_A(j)}{M_j}$ \Comment{Mean for ID $j$}
            \State $ES_A(y) \gets ES_A(y) + IDS_A(j)$
            \State $K_y \gets K_y + 1$
        \EndFor
        \State $ES_A(y) \gets \frac{ES_A(y)}{K_y}$ \Comment{Mean for Ethnicity $y$}
    \EndFor

    \State $y^* \gets \arg\min_{y \in \mathcal{E}} ES_A(y)$ \Comment{Ethnicity with $\downarrow$ score}
    \State $j^* \gets \arg\min_{j \in y^*} IDS_A(j)$ \Comment{ID with $\downarrow$ score}
    \State \textbf{Remove} $j^*$ from $y^*$
    \State $N \gets N + 1$
\EndWhile
\end{algorithmic}
\end{algorithm}

\subsubsection{Protocol B}

Protocol A relies on two strong assumptions: 1) The number of identities in an ethnicity is not relevant for balancing; 2) The number of images per identity is not relevant too. Although these assumptions seem to hold sufficiently well in practice, it can be noticed that retaining the maximum performance requires a relaxation of the second assumption. For this, we theorize that a sampling approach that considers, not the average score of all images, but the sum of the score of all images leads to more robust solutions. With this change, the choice between two identical identities, regarding their ethnicity score, relies on their number of images. For this, Protocol B, $IDS_{Bj}$ is given by
\vspace{-0.15cm}
\begin{align}
     IDS_{Bj} = \sum^{M_j}_i ies_{j,i}
\end{align}

This formulation incorporates information regarding the number of images, preferring images that maximise both the image score and the number of images. So, between two ethnically similar identities, the one with the lowest amount of images is more likely to be discarded. This allows for a greater performance retention, while still providing a strong bias reduction scheme. The final score is calculated as
\vspace{-0.15cm}
\begin{align}
     ES_{By} = \sum^{N_y}_j \frac{IDS_{Bj}}{N_y}
\end{align}

Despite considering the number of images within an identity, the ethnicity score ignores the number of identities contained in the ethnicity group. The elimination of identities follows a similar process to Protocol A as shown in Algorithm~\ref{alg:protocolB}.

\begin{algorithm}
\caption{Sampling Process for Protocol B}\label{alg:protocolB}
\begin{algorithmic}
\Require $Z \geq 0$
\State $\mathcal{E} \gets \{Af, As, Ca, In\}$ \Comment{Ethnic Groups: African, Asian, Caucasian, Indian}
\State $N \gets 0$
\While{$N \neq Z$}
    \For{$y \in \mathcal{E}$}
        \State $ES_B(y) \gets 0$  \Comment{Initialize ethnicity-level score}
        \State $K_y \gets 0$ \Comment{Ethnicity count for sample $j$}
        \For{$j \in y$} 
            \State $IDS_B(j) \gets 0$ \Comment{Initialize ID-level score}
            \For{$img \in j$}
                \State $IDS_B(j) \gets IDS_B(j) + ies(img)$
            \EndFor
            \State $ES_B(y) \gets ES_B(y) + IDS_B(j)$
            \State $K_y \gets K_y + 1$
        \EndFor
        \State $ES_B(y) \gets \frac{ES_B(y)}{K_y}$ \Comment{Mean for Ethnicity $y$}
    \EndFor

    \State $y^* \gets \arg\min_{y \in \mathcal{E}} ES_B(y)$ \Comment{Ethnicity with $\downarrow$ score}
    \State $j^* \gets \arg\min_{j \in y^*} IDS_B(j)$ \Comment{ID with $\downarrow$ score}
    \State \textbf{Remove} $j^*$ from $y^*$
    \State $N \gets N + 1$
\EndWhile
\end{algorithmic}
\end{algorithm}

\subsubsection{Protocol C}

Similarly to the relaxation on the assumptions of Protocol A to construct Protocol B, one can further relax the first assumption regarding the independence between fairness and performance metrics, and the number of identities per ethnicity. Hence, we perform a similar relaxation step on the computation of the ethnicity score. Besides this relaxation, one strong benefit of this hypothesis is the different sampling strategy, which supports extreme identity removal strategies. In theory, such method could function until there is a single identity per ethnicity, creating a robust approach for highly imbalanced datasets. In Protocol C, $IDS_{Cj}$ is computed as 
\vspace{-0.2cm}

\begin{align}
     IDS_{Cj} = \sum^{M_j}_i ies_{j,i},
\end{align}

in a similar approach to $IDS_{B}$, hence, considering the number of images within an identity. $ES_{Cy}$ is given by
\vspace{-0.2cm}

\begin{align}
     ES_{Cy} = \sum^{N_y}_j IDS_{Cj}
\end{align}

This score considers both the number of images in an identity, propagated from $IDS_{Cj}$, and the number of different identities belonging to a demographic group $y$. This formulation as a sum over the different identities implies a different strategy to select which identity should be removed.  On Protocols A and B, the final score is computed as a mean over all identities, so if we remove the one with the lowest ethnicity score, the average score increases, thus approximating that group to other high-scoring ethnicity groups. In this case, an identity is removed from the group $y$ with the highest $diag(ES_C)^{(y)}$, hence, pushing it down to the level of the other groups. This is highlighted in Algorithm~\ref{alg:protocolC}.

\begin{algorithm}
\caption{Sampling Process for Protocol C}\label{alg:protocolC}
\begin{algorithmic}
\Require $Z \geq 0$
\State $\mathcal{E} \gets \{Af, As, Ca, In\}$ \Comment{Ethnic Groups: African, Asian, Caucasian, Indian}
\State $N \gets 0$
\While{$N \neq Z$}
    \For{$y \in \mathcal{E}$}
        \State $ES_C(y) \gets 0$  \Comment{Initialize ethnicity-level score}
        \For{$j \in y$} 
            \State $IDS_C(j) \gets 0$ \Comment{Initialize ID-level score}
            \For{$img \in j$}
                \State $IDS_C(j) \gets IDS_C(j) + ies(img)$
            \EndFor
            \State $ES_C(y) \gets ES_C(y) + IDS_C(j)$
        \EndFor
    \EndFor

    \State $y^* \gets \arg\max_{y \in \mathcal{E}} ES_C(y)$ \Comment{Ethnicity with $\uparrow$ score}
    \State $j^* \gets \arg\min_{j \in y^*} IDS_C(j)$ \Comment{ID with $\downarrow$ score}
    \State \textbf{Remove} $j^*$ from $y^*$
    \State $N \gets N + 1$
\EndWhile
\end{algorithmic}
\end{algorithm}

\subsubsection{Relabeling}

Low-scored samples might be an indicative of the presence of mislabelled samples. In fact, their presence might pose the question regarding the potential performance of face recognition models on properly labelled datasets. Hence, given a dataset we propose the creation of a second version of that dataset with adjusted labels, determined by an ethnicity score agent. This relabelled dataset provides an alternative starting point for our balancing protocols, enabling us to assess the impact of mislabelling on dataset balancing.
In our relabelling process, each identity’s ethnicity label $y$ is reassigned based on the maximum score inferred by the ethnicity score agent: $y = argmax_{\hat{y}\in \mathcal{D}} IDS^{(\hat{y})}_{Aj}$. Identities that were considered low-score in one dataset can become high-score following their ethnicity label re-assignment, thus reducing their likelihood of being removed. Our theory shall hold even when these noisy signals are removed through the relabelling process. Thus, by replicating protocols on both datasets, we show that balancing goes beyond a simple strategy to remove mislabelled samples.  

% \textcolor{red}{MISSING A BAR PLOT WITH THE SCORE FOR ALL METHODS OVER SAMPLING. NOW WE ONLY HAVE FOR THE IDS.}

% \begin{itemize}
%     \item A - 
%         \begin{equation}
%             \frac{1}{N_i}\sum_{j=1}^{N_i}{\sum_{k=1}^{M_j}{\frac{es_{j,k}}{M_j}}}
%         \end{equation}
    
%     \item D - 
%         \begin{equation}
%             \frac{1}{N_i}\sum_{j=1}^{N_i}{\sum_{k=1}^{M_j}{es_{j,k}}}
%         \end{equation}
    
%     \item E - 
%     \begin{equation}
%         \sum_{j=1}^{N_i}{\sum_{k=1}^{M_j}{es_{j,k}}}
%     \end{equation}

% \end{itemize}

\section{Experimental setup}
\label{sec:exp}

In this section, we expand on the details regarding the experiments conducted and the reproducibility of the methods previously described. All sampled versions, the code for sampling and other elements will be publicly available on GitHub\footnote{\url{https://github.com/NetoPedro/Equilibrium-Face-Recognition}}.

\subsection{Datasets}

Provided our hypothesis, and our methodology to test it, one can leverage datasets that contain, or not, ethnicity information. However, for a comparison between a discretely balanced dataset and a continuously balanced one, it was essential to select a dataset that was discretely balanced with regards to ethnicity. For this, BUPT-BalancedFace~\cite{wang2021meta} was chosen as it is naturally balanced for a same number of identities per ethnicity. BUPT-GlobalFace~\cite{wang2021meta} was used for extended experiments which started from an unbalanced starting point, and evaluation was conducted on a ethnicity-balanced dataset, known as RFW~\cite{Wang2019RFW}, to assess fairness metrics. These datasets are described in the following subsections.

\subsubsection{RFW}

Racial Faces in-the-wild (RFW)~\cite{Wang2019RFW} has been designed to benchmark face recognition algorithms on the different ethnicity groups. Hence, it is divided, by the original authors, into four groups, each corresponding to an ethnicity. Each of these groups has roughly 6000 images, and in total the dataset is composed of approximately 9000 different individuals. It should be noted that in face recognition scenarios, the presence of bias is measured intra-ethnicity. In other words, if pairs of ethnicity A are easier to classify than pairs of ethnicity B, then our model is biased against these ethnicities. 

\begin{figure*}[t!]
    \centering
    \begin{subfigure}[t]{0.325\linewidth}
        \centering
        \includegraphics[width=\linewidth]{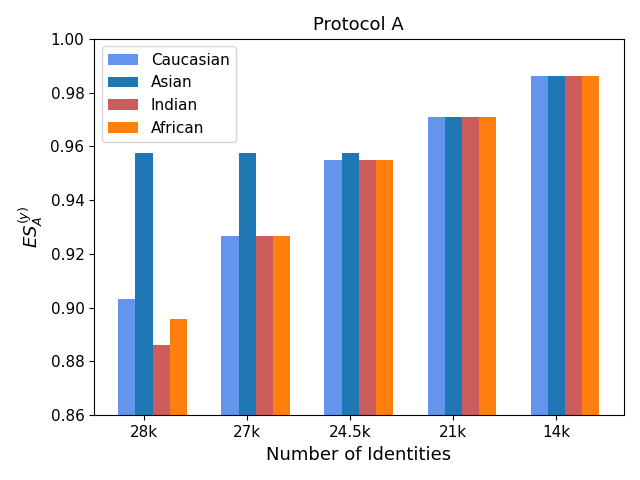}
    \end{subfigure}%
    ~
    \begin{subfigure}[t]{0.325\linewidth}
        \centering
        \includegraphics[width=\linewidth]{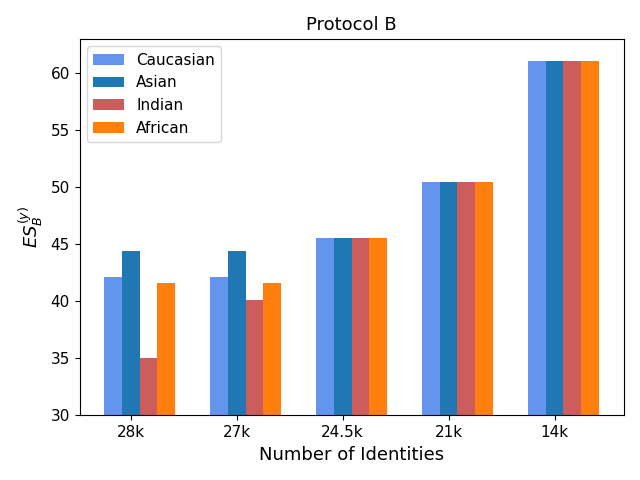}
    \end{subfigure}
    \begin{subfigure}[t]{0.325\linewidth}
        \centering
        \includegraphics[width=\linewidth]{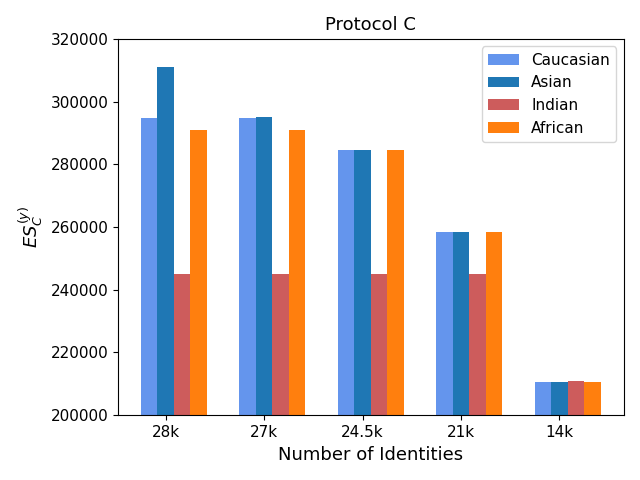}
    \end{subfigure}
    \caption{The evolution of the different scores $diag(ES_A)^{(y)}$, $diag(ES_B)^{(y)}$, and $diag(ES_C)^{(y)}$ (y-axis) as the number of identities in the dataset decreases (x-axis). It can be seen that for the first two protocols, balance is quickly achieved, whereas for the other only the last iteration is balanced. }
    \label{fig:sampled_ethnicity_score_evolution}
\end{figure*}

\subsubsection{BUPT-BalancedFace and BUPT-GlobalFace}

Both BUPT-BalancedFace and BUPT-GlobalFace~\textit{et al.}~\cite{wang2021meta} have been collected to be used as training datasets for fair face recognition models. Proposed by the same authors of RFW, these two datasets can be used to study the bias inherent to the dataset and to the architecture itself. Similar to RFW, each identity has a label based on its skin tone and is further categorised into one of the following ethnicities: African, Asian, Indian or Caucasian. BUPT-GlobalFace is the largest of the two datasets, with two million images from 38k identities. The ethnicity distribution of these identities follows a similar distribution to the world's population. BUPT-BalancedFace equalises the number of identities that belong to each ethnicity, and is composed of 1.3 million images from 28k distinct identities. This dataset is balanced with respect to its discrete labels.

\subsection{Balancing Criteria}

Different protocols reach their equilibrium at different points, for instance, some might originate a balanced dataset after removing three thousand identities, others might require a higher number of removals. Figure~\ref{fig:sampled_ethnicity_score_evolution} shows the equilibrium of the different ethnicities for different numbers of identities for all the three protocols. As expected, Protocol A and B reach even scores around a similar number of identities, while Protocol C requires significantly more steps to balance all four ethnicities.

\begin{figure}[h!]
    \centering
    \includegraphics[width=1.05\linewidth]{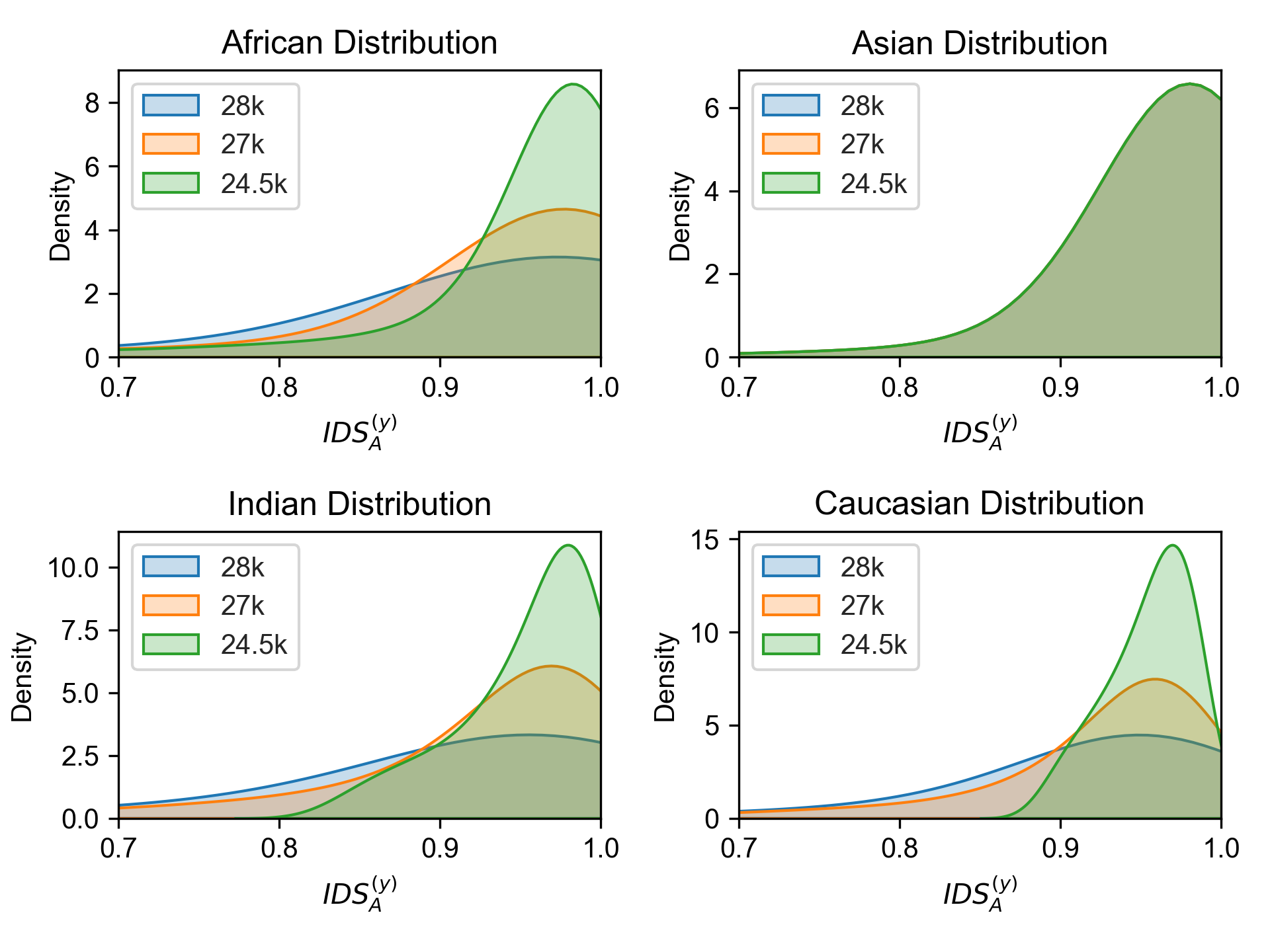}
    \caption{Distribution shift of the ethnicity score on the different ethnicity groups when sampling identities with Protocol A. It can be seen that the Asian distribution is not affected, as no sample is removed. It would only be affected when the sampling removes up to 25\% of the identities. Additionally, this shows that distributions shift from a flatter look into one that peaks near 1. }
    \label{fig:sampling_diff_ethnicities_effect}
\end{figure}

A fair comparison of all these protocols requires them to have a similar number of identities. As such, within the context of this work, we have decided to compare the different approaches through the removal of a fixed number of identities. We compare the original dataset with 28k identities with sampled versions with 27k, 24.5k, 21k and 14k, thus removing from 3.5\% to 50\% of the identities. One should note that this are reference values for comparison, and might not necessarily reflect the optimal value for identity removal. In Figure~\ref{fig:sampling_diff_ethnicities_effect} we show a density approximation of Protocol A identity scores computed from the available identities. Once we remove 1k and 3.5k identities, the mean of these densities, for three ethnicities, starts to converge towards to 1, and the variance starts to decrease. In addition, we can see that $IDS_A^{(y)}$ shows no change as we remove up to 3.5k identities from the dataset, reflecting the behaviour of this protocol, which only affects this demographic group for higher values of identities removed. This results from the higher baseline $IDS_A^{(y)}$ for this group. 

\subsection{Ethnicity Score as an Utility Measure}

The nature of this score and its link to performance improvements raises the question of whether it acts as a utility measure. In face recognition, utility-like scores have been quite popular due to their ability to give higher or lower importance to images given their utility for training a face recognition system. Some of these utility measures are linked to specific models, others are more agnostic. In practice, our ethnicity score behaves like an agnostic utility system, providing insights on the most impactful samples per ethnicity.

Due to the similarities with utility, one might pose the question regarding how our score differs from utility scores, and if they can be combined. For this assessment, we performed a comparison between $ies_{j,i}^{(y)}$ and a quality score. For this, we compared our score with the quality score produced by CR-FIQA~\cite{DBLP:conf/cvpr/BoutrosFKFD23}.

% \begin{figure}
%     \centering
%     \includegraphics[width=\linewidth]{qualityvsethnicity.jpg}
%     \caption{Relationship between quality and ethnicity score. There seems to be no relationship between the two. \textcolor{red}{Change x-axis label}}
%     \label{fig:quality}
% \end{figure}

\begin{figure}[t!]
    \centering
    \begin{subfigure}[t]{0.495\linewidth}
        \centering
        \includegraphics[width=\linewidth]{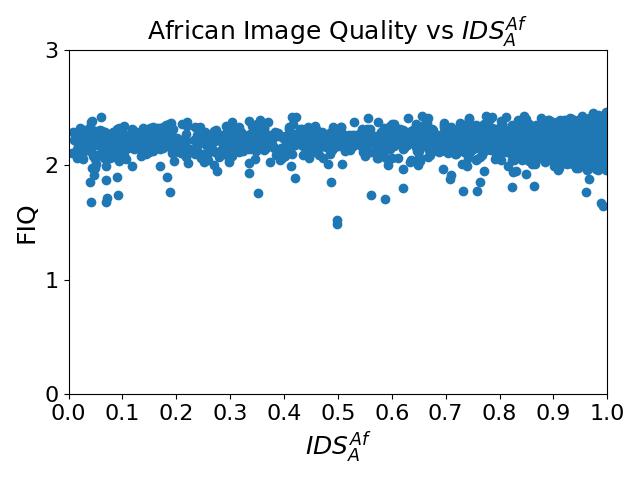}
    \end{subfigure}%
    \begin{subfigure}[t]{0.495\linewidth}
        \centering
        \includegraphics[width=\linewidth]{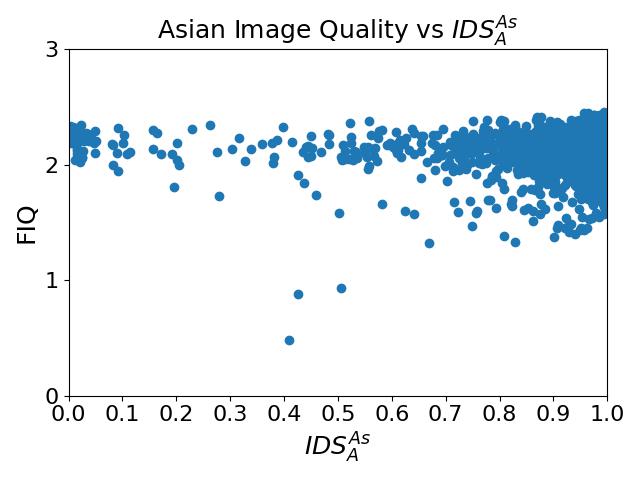}
    \end{subfigure}
    \begin{subfigure}[t]{0.495\linewidth}
        \centering
        \includegraphics[width=\linewidth]{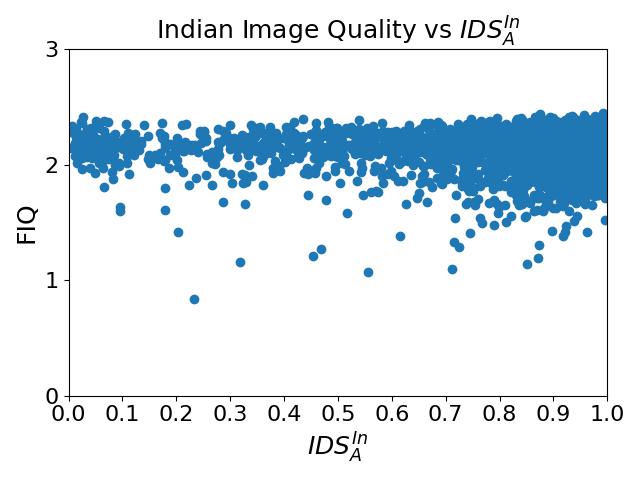}
    \end{subfigure}
    \begin{subfigure}[t]{0.495\linewidth}
        \centering
        \includegraphics[width=\linewidth]{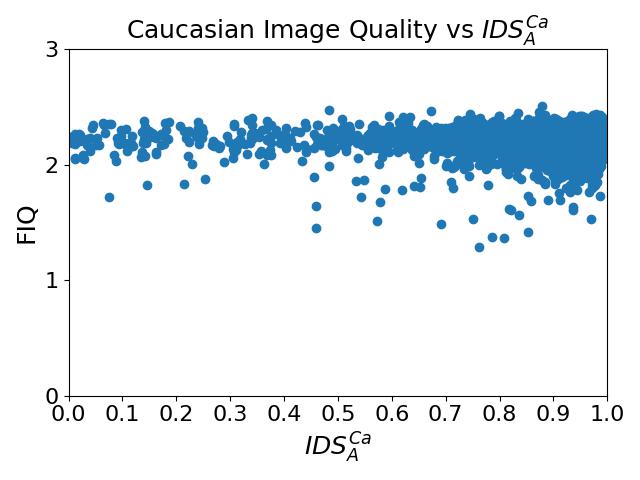}
    \end{subfigure}
    \caption{}
    \label{fig:quality}
\end{figure}

Figure~\ref{fig:quality} highlights that high-scores for the quality metric do not imply high scores on our ethnicity score. Furthermore, it can be seen that samples with maximum representativeness of their ethnicity group might still have subpar quality scores in comparison with less representative samples. Hence, despite the visible differences in the quality distribution of different ethnicities (e.g. $IDS_A^{In}$ shows higher dispersion on the y-axis), samples removed by our balancing strategy are not necessarily low-utility ones. As such, the ethnicity score can be seen as an uncorrelated utility metric when compared to quality-based metrics, indicating the potential of combining both metrics for the selection of fair and high-performing datasets. 

\subsection{Ethnicity Classifier}

Both training datasets have ethnicity information in the form of discrete labels. To study the impact of continuous labels, there is no dataset available in the literature that contains such information. Hence, we have designed a pseudo-labelling strategy to infer such information. Starting from a pretrained face recognition model (ElasticArc~\cite{boutros2022ElasticFace} with a ResNet100~\cite{he2016deep}), we add a classification layer with four output classes, one for each ethnicity. Despite believing that the distance between African and Caucasian samples, and between Caucasian and Indian samples is not equal, an ordinal formulation of this classification problem would be an underfitted formulation of those distances. Hence, we train our classification model on BUPT-BalancedFace with each ethnicity hot-encoded in a vector. During training, only the weights of the classification layer are updated. We verified that the latent space of FR systems encodes ethnicity information, as this approach achieves far stronger results than training a deep neural network from scratch.

\begin{figure*}[t!]
    \centering
    \begin{subfigure}[t]{0.485\linewidth}
        \centering
        \includegraphics[width=\linewidth]{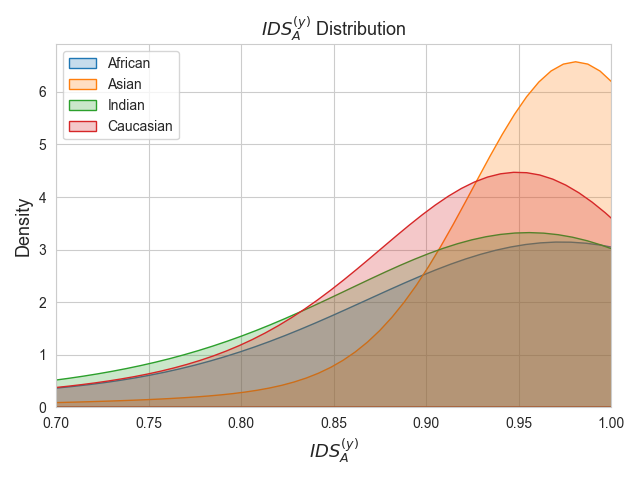}
        \caption{Original Distributions}
        \label{fig:ethnicity_score_distribution}
    \end{subfigure}%
    ~
    \begin{subfigure}[t]{0.485\linewidth}
        \centering
        \includegraphics[width=\linewidth]{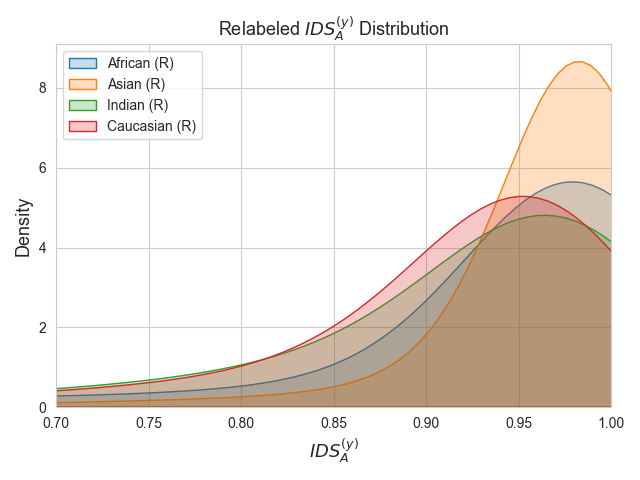}
        \caption{Relabeled Distributions}
        \label{fig:ethnicity_score_distribution_relabel}
    \end{subfigure}

    \caption{Identity Score ($IDS^{(y)}_A$) Distribution for the different demographic groups for both the original ethnicity labels and after a relabelling process. }
    \label{fig:sampled_ethnicity_score_evolution}
\end{figure*}

% \begin{figure}
%     \centering
%     \includegraphics[width=0.8\linewidth]{ethnicity_base_distribution.png}
%     \caption{\textbf{Ethnicity Score (ES) Distribution} for the different demographic groups described in the original labels. }
%     \label{fig:ethnicity_score_distribution}
% \end{figure}

% \begin{figure}
%     \centering
%     \includegraphics[width=0.8\linewidth]{ethnicity_relabeled_distribution.png}
%     \caption{\textbf{Relabeled Ethnicity Score (ES) Distribution} for the different demographic groups constructed from the score itself.}
%     \label{fig:ethnicity_score_distribution_relabel}
% \end{figure}

After training our classifier, we use it to infer a probability vector for each image in both BUPT-BalancedFace and BUPT-GlobalFace datasets. One important point is to take into account that an image with a high probability on the Caucasian group is not necessarily representative of the definition of Caucasian, it means that that particular image is easily classified as belonging to that group of images in the dataset. Given this, we can also verify that images with the lowest probability are likely to not contribute significantly to the performance of that demographic group. Figure~\ref{fig:ethnicity_score_distribution} shows, for each group, a density function of our proposed score, further described in the following sections. Details on the construction of this score are given in the following section. In the image, it is visible that the distribution of these scores in different groups varies significantly. This can be mitigated, if we relabel the ethnicity label of each identity with the new ES, as seen in Figure~\ref{fig:ethnicity_score_distribution_relabel}.  

\subsection{Implementation Details}

In our experiments, we utilise three different architectures: iResNet-34, -50 and -100~\cite{he2016deep}, which output an embedding of 512-D, as these have been widely used in the literature of fair face recognition~\cite{Neto2023CompressedModels,wang2021meta,neto2024knowledge,caldeira2024mst,deandres2024frcsyn}. The embedding size is standard across the literature~\cite{deng2022ArcFace,boutros2022ElasticFace}, and we utilise iResNet-34 for the majority of the experiments, as it is faster to train and validate the effects of different training datasets. Aligned with previous research~\cite{deng2022ArcFace,caldeira2024mst,wang2021meta}, all networks have been trained from scratch without resorting to any pre-trained weights. We optimise the networks with margin-based losses, and for a more robust evaluation, we conduct experiments with two different losses: ElasticArcFace~\cite{boutros2022ElasticFace} and CurricularFace~\cite{huang2020Curricular}. Both losses use the same value for the margin, $m = 0.50$, and for the scaling factor, $s=64$. In addition, ElasticArcFace has a standard deviation parameter set to 0.05 as described in the original paper. All these hyper-parameters have been defined as given by their original implementation.

Each network is optimised in batches of 256 images for 26 epochs, with an initial learning rate of 0.1. We schedule the learning rate of the stochastic gradient descent (SGD) optimiser to decay at epochs 8, 14, 20 and 25. Random horizontal flips are the only augmentation strategy used. These elements replicate what has been done in the literature~\cite{deng2022ArcFace, boutros2022ElasticFace}.

\subsection{Performance Evaluation}

To properly evaluate the different sampling approaches and to compare models trained with them with other models, we followed a strategy of comparing datasets with the same number of identities. So, having three proposed sampling approaches, we wanted to compare them with a fourth model trained on an approach that retained the balance of the dataset with respect to the original discrete labels. Hence, we perform random sampling, which will retain the discretely balanced nature of the original set. So, for each sampling strategy, we compare a model trained on its respective dataset with models trained on the original dataset and a randomly sampled dataset. The first comparison highlights if there is any performance and fairness change, whereas the second shows the advantages of continuous labels against discrete labels on equally sized datasets.

To evaluate both performance and fairness, we use RFW and individually compute the accuracy for each demographic group. This accuracy across demographic groups can be used to compute a global accuracy, which showcases the discriminative power of the final model. Additionally as seen in previous works~\cite{mamede2024fairness,wang2021meta,deandres2024frcsyn,neto2024knowledge}, the standard-deviation (STD) across the accuracy of different groups can be measured as a guide to its fairness. The skewed error rate (SER) can further enhance our understanding of the relation between the higher and lower error rates within the four ethnicities. SER can be described, with respect to the accuracy, as follows:

\vspace{-0.15cm}
\begin{align}
     SER =  \frac{1 - min([Acc_{af},Acc_{as},Acc_{ca},Acc_{in}])}{1 - max([Acc_{af},Acc_{as},Acc_{ca},Acc_{in}])}
\end{align}

The combination of performance and fairness metrics ensures that we aim for the best-performing model at a specific bias level. In other words, reducing the bias at a high performance cost can increase the errors across all ethnicity. For this, we further analyse this trade-off through the study of the Pareto-Front of all models with respect to prediction error and bias.   

\subsection{Hardware}
All face recognition models and the ethnicity classifier have been trained on a server with a single NVIDIA A100 80GB GPU. The sampled versions of the datasets have been generated on a commercial Apple's MacBook Pro with a M1 Pro chip.

\begin{table}[h!]
\footnotesize
\renewcommand{\arraystretch}{1.0}
\setlength{\tabcolsep}{7pt}
\centering
\caption{Results for an iResNet-34 (ElasticArcFace) when half of the samples from a single identity are removed. Three strategies are followed: Retaining 3500 random identities (Rand), the 3500 identities that minimise the score for that ethnicity (Min), and the 3500 identities that maximise this score (Max).   }
\label{tab:single_ethnicity_sampling}
\begin{tabular}{lllrrr}
\Xhline{2\arrayrulewidth}
      \#ID  & Strategy   &    \multicolumn{4}{c}{Accuracy per skin tone (\%) $\uparrow$}  \\
           \cmidrule(rl){3-6}
   & & Caucasian      & \multicolumn{1}{c}{Indian} &\multicolumn{1}{c}{Asian} & \multicolumn{1}{c}{African} \\
\hline
28k  & None &  96.67\%  & 94.88\% & 94.22\% & 93.38\%      \\
\hline
\hline
 24.5k & Rand. (Af) &  96.08\% & 94.28\%  & 93.82\% & 91.92\% \\
\hline
 24.5k & Min. (Af) & \textbf{96.30\%} & 94.52\% & \textbf{94.37\%} &  89.95\%  \\
\hline
 24.5k & Max. (Af) & 96.12\%  & \textbf{94.77\%}  & 94.18\% & \textbf{92.78\%}  \\
\hline
\hline
 24.5k & Rand. (In) &  96.30\% & 94.02\%  & 93.85\% & \textbf{93.38\%}  \\
 \hline
 24.5k & Min. (In) & 96.35\%  & 93.78\%  & \textbf{94.23\%} & 93.15\%  \\
\hline
 24.5k & Max. (In) & \textbf{96.38\%} & \textbf{94.68\%} & 93.98\% &  93.13\% \\
\hline
\hline
 24.5k & Rand. (As) &  96.43\% & \textbf{94.67\%}  & 92.95\% & 93.45\% \\
\hline
 24.5k & Min. (As) & 96.28\% & 94.42\% & 	\textbf{93.00\%} &  93.30\%  \\
\hline
 24.5k & Max. (As) & \textbf{96.46\%}  & 94.52\%  & \textbf{93.00\%} & \textbf{93.85\%}  \\
\hline
\hline
 24.5k & Rand. (Ca) &  95.60\% & 94.63\%  & 93.80\% & 93.48\% \\
\hline
 24.5k & Min. (Ca) & 95.70\% & 94.72\% & \textbf{94.07\%} &  93.43\%  \\
\hline
 24.5k & Max. (Ca) & \textbf{95.77\%}  & \textbf{94.97\%}  & 93.97\% & \textbf{93.52\%}  \\
\hline
\hline
\Xhline{2\arrayrulewidth}
\end{tabular}
\end{table}

\begin{table*}[ht]
\renewcommand{\arraystretch}{1.0}
\setlength{\tabcolsep}{7pt}
\centering
\caption{Results for iResNet-34 (ElasticArcFace) trained on several versions of the training set. Ranging from no sampling to three different sampling strategies, where two are based on Protocol A - \textbf{A} and \textbf{A(R)}. These latter strategies seem to hold significant advantages with respect to random sampling. Results are shown for several variations of the number of identities kept in the training set.  }
\label{tab:protocolA_elastic}
\begin{tabular}{lllrrrrrr}
\Xhline{2\arrayrulewidth}
      \#ID  & Strategy   &    \multicolumn{4}{c}{Accuracy per skin tone (\%) $\uparrow$} & Average &\multicolumn{2}{c}{Fairness $\downarrow$}\\
           \cmidrule(rl){3-6}\cmidrule(rl){8-9}  
   & & Caucasian      & \multicolumn{1}{c}{Indian} &\multicolumn{1}{c}{Asian} & \multicolumn{1}{c}{African} &  & \multicolumn{1}{c}{STD} & \multicolumn{1}{c}{SER}    \\
\hline
28k  & None &  96.67\%  & 94.88\% & 94.22\% & 93.38\%  & 94.79\%& 1.39 & 1.99     \\
\hline
\hline
27k & Random & 96.65\%  & 94.82\%  & 94.20\% & 93.77\% & 94.86\% & 1.27 & 1.86 \\
\hline
24.5k & Random & 96.28\%  & 94.47\%  & 93.55\% & 93.15\% & 94.36\% & 1.39 & 1.84  \\
\hline
21k & Random & 95.78\%  & 93.97\%  & 93.20\% & 92.43\% & 93.85\% & 1.44	 & 1.79 \\
\hline
14k & Random & 93.82\%  & 92.10\%  & 91.08\% & 90.00\% & 91.75\% & 1.62 & 1.62 \\
\hline
\hline
27k & A & 96.48\%  & 94.90\%  & 94.33\% & 93.77\% & \textbf{94.87}\% & \textbf{1.17} & \textbf{1.76} \\
\hline
%26.25k & A & 96.43\%  & 95.05\%  & 94.07\% & 93.83\% & 94.85\% & 1.18  &  1.72 \\
%\hline
24.5k & A & 96.11\%  & 94.82\%  & 94.23\% & 93.42\% & \textbf{94.65}\% & 1.13 & 1.69  \\
\hline
21k & A & 95.50\%  & 93.92\%  & 93.78\% & 92.82\% & \textbf{94.01}\% & 1.11 &  1.59 \\
\hline
14k & A & 92.01\%  & 91.83\%  & 92.20\% & 90.95\% & 91.75\% & \textbf{0.55}  & \textbf{1.16}  \\
\hline
\hline
27k  & A(R) &96.35\%  & 94.80\%  & 94.30\% & 93.58\% & 94.76\% & \textbf{1.17} & \textbf{1.76} \\
\hline
%26.25k & B&96.48\%  & 95.03\%  & 94.23\% & 93.30\% & 94.76\% & 1.35 & 1.90 \\
%\hline
24.5k & A(R) & 96.00\%  & 94.37\%  & 94.13\% & 93.52\% & 94.51\% &  \textbf{1.06} & \textbf{1.62} \\
\hline
21k & A(R) & 95.13\%  & 93.87\%  & 93.63\% & 93.03\% & 93.92\% & \textbf{0.88} & \textbf{1.43} \\
\hline
14k & A(R) &92.22\%  & 92.42\%  & 92.12\% & 90.68\% & \textbf{91.86}\% & 0.80 & 1.23 \\
\hline
\hline
\Xhline{2\arrayrulewidth}
\end{tabular}
\end{table*}

\section{Results}
\label{sec:results}

Our hypothesis that a continuos label is essential to ensure and measure the balance nature of a dataset can be further validated through experimental results. In this following section, we focus on validating the previously discussed assumptions that shall hold for our hypothesis to be verified. Hence, the following section discusses the validation of the two main assumptions: 1) If continuos labels have an impact removing most representative samples has a different effect from removing the less representative, or randomly; 2) Regardless of the Protocol we follow for balancing the data  in the continuos space (through sample removal), it shall attain fairer results than a discretely balanced dataset with the same number of identities.

All the findings presented in this section originate from experiments using an iResNet-34 with ElasticArcFace. To showcase the germinability of our hypothesis beyond network architecture or loss, experiments with other architectures, losses, and training datasets can be found in the supplementary material. 

\subsection{Single Ethnicity Analysis}

In our hypothesis, the first assumption is that not all samples contribute equally to the performance of the model on a given ethnicity. Hence, some might be deemed more representative, or less representative of a given ethnicity. Given this assumption, one can estimate that retaining less representative samples of one ethnicity would lead to lower performance on that specific ethnicity than retaining the most representative ones. This shall be more noticeable on demographic groups frequently associated with unfairness. 

To validate such assumption, we propose to remove exactly half of the identities in a given ethnicity. To ensure that less representative samples lead to lower performance on that ethnicity when compared to more representative ones, we propose \textbf{Min.} and \textbf{Max.}, the retention of the less and the most representative half, respectively. To add an additional layer of granularity, we also state that randomly (\textbf{Rand.}) removing half of the identities shall lead to lower performance too. 

As expected, Table~\ref{tab:single_ethnicity_sampling} validates our assumption that training with the most representative samples of an ethnicity (\textbf{Max.}) leads to higher performance on that given ethnicity. This is strongly noticeable on the African group with nearly three percent point difference. Max. for both African and Indian groups retains performances comparable to a model trained on the entire dataset.  An additional observation is that despite training on significantly less data, some versions seem to improve, on certain ethnicities, when compared to a model trained on the entire dataset. Fairness is not considered in this specific analysis, as the focus was not to balance the data, but to validate the first assumption that not all identities of a given ethnicity contribute equally to that given ethnicity.

\begin{table*}[ht]
\renewcommand{\arraystretch}{1.0}
\setlength{\tabcolsep}{7pt}
\centering
\caption{ Results for iResNet-34 (ElasticArcFace) trained on several versions of the training set. Ranging from no sampling to three different sampling strategies, where two are based on Protocol B - \textbf{B} and \textbf{B(R)}. These latter strategies seem to hold significant advantages with respect to random sampling. Results shown for several variations of the number of identities kept in the training set. 
}
\label{tab:protocolb_elastic}
\begin{tabular}{lllrrrrrr}
\Xhline{2\arrayrulewidth}
      \#ID  & Strategy   &    \multicolumn{4}{c}{Accuracy per skin tone (\%) $\uparrow$} & Average &\multicolumn{2}{c}{Fairness $\downarrow$}\\
           \cmidrule(rl){3-6}\cmidrule(rl){8-9}  
   & & Caucasian      & \multicolumn{1}{c}{Indian} &\multicolumn{1}{c}{Asian} & \multicolumn{1}{c}{African} &  & \multicolumn{1}{c}{STD} & \multicolumn{1}{c}{SER}    \\
\hline
28k  & None &  96.67\%  & 94.88\% & 94.22\% & 93.38\%  & 94.79\%& 1.39 & 1.99     \\
\hline
\hline
27k & Random & 96.65\%  & 94.82\%  & 94.20\% & 93.77\% & 94.86\% & 1.27 & 1.86 \\
\hline
24.5k & Random & 96.28\%  & 94.47\%  & 93.55\% & 93.15\% & 94.36\% & 1.39 & 1.84  \\
\hline
21k & Random & 95.78\%  & 93.97\%  & 93.20\% & 92.43\% & 93.85\% & 1.44	 & 1.79 \\
\hline
14k & Random & 93.82\%  & 92.10\%  & 91.08\% & 90.00\% & 91.75\% & 1.62 & \textbf{1.62} \\
\hline
\hline
27k & B& 96.60\%  & 95.20\%  & 94.45\% & 	93.77\% & \textbf{95.02}\% & 1.24 & 1.87 \\
\hline
24.5k & B&96.18\%  & 94.73\%  & 94.32\% & 93.98\% & \textbf{94.80\%} & \textbf{0.97}&  \textbf{1.58} \\
\hline
21k & B& 96.27\%  & 94.70\%  & 94.25\% & 93.68\% & \textbf{94.73\%} & \textbf{1.11} &  \textbf{1.69} \\
\hline
14k & B& 95.03\%  & 93.18\%  & 92.42\% & 91.22\% & 92.96\% & 1.60 & 1.77 \\
\hline
\hline
27k & B(R)& 96.37\%  & 95.08\%  & 94.42\% & 93.73\% & 94.90\% & \textbf{1.12}  & \textbf{1.73} \\
\hline
24.5k & B(R)& 96.25\%  & 94.78\%  & 94.33\% & 93.65\% & 94.75\% & 1.10  & 1.69 \\
\hline
21k & B(R)& 96.10\%  & 94.38\%  & 94.07\% & 93.13\% & 94.42\% & 1.24  & 1.76 \\
\hline
14k & B(R)& 95.10\%  & 93.70\%  & 92.38\% & 91.83\% & \textbf{93.25\%} & \textbf{1.46}  & 1.67 \\
\hline
\hline
\Xhline{2\arrayrulewidth}
\end{tabular}
\end{table*}

\subsection{Balancing in the Continuous Space}

Following the verification of the first assumption, we now are assured that the representativeness of the identity is, to some extent, represented by the continuous label. Hence, we now have an additional assumption that must be held if one wants to validate our hypothesis. Balancing the data in this new continuous space must lead to superior performance vs. fairness trade-offs. 

The concept of balancing is, in itself, a vague concept, which might refer to balanced with regards to the identities, to the density of each identity, or even just to the raw number of images. And thus, to show the clearly superior performance of our hypothesis, leaving no question regarding the continuous nature of the problem, we show independently that our assumption holds for these three distinct perspectives of a balanced dataset.

\subsubsection{Protocol A}

In the first scenario, one does not account for the density of each identity nor to the number of identities left in an ethnicity group. Rather, one uses Protocol A, which only considers our proposed label.

On Table~\ref{tab:protocolA_elastic} three distinct approaches for sampling are used. 1) A fixed amount of identities are sampled randomly, hence, retaining the balancing of the data in the discrete space - \textbf{Random} ; 2) using the original ethnicity labels and the proposed score, a fixed amount of identities are removed - \textbf{A}; 3) similar to the latter, it utilises the revised discrete labels and the proposed score - \textbf{A(R)}. Given our initial assumption that fairness metrics shall improve under our strategies, we found that beyond fairness metrics, balancing the data in the continuous space consistently led to improved performance metrics too. Hence, not only was the assumption verified, but it also showed that continuous labels have such an impact on data balancing that even without accounting for performance-related factors (\#IDs and \#images), performance can be improved with correct balancing.

\subsubsection{Protocol B}

Differently from the strategy previously discussed, Protocol B leverages continuous labels, but also accounts for the number of images within an identity. Through this, we add an additional focus on the exploitation of the trade-off between fairness and performance. To validate our assumption regarding balancing in the continuous space, this protocol would have to show reduced fairness, and improved performance. We consider two different approaches for this protocol. One with a revision of the original labels, \textbf{B(R)}, and one with the original ethnicity labels, \textbf{B}.

Table~\ref{tab:protocolb_elastic} highlights some of the relevant aspects of exploring continuous labels in line with the number of images per identity. First, although having slightly higher bias than Protocol A in some iterations, it is always better than the bias present in all versions of random sampling. Additionally, the average accuracy is greatly improved compared to both protocol A and Random. When we remove 1k of the identities, the performance increases significantly with improvements on both STD and SER. But the most surprising result is that after removing one quarter of the dataset (7k identities), the performance degradation is nearly non-existent (0.06 percent points) with a high improvement on the bias. This means that we managed to train a comparable model with respect to performance, which is more fair and takes roughly 25\% less time to train. When we remove half of the identities, the proposed B(R) achieves a performance 1.5 percent points above the random equivalent, while also being at 1.5 percent points distance from the version trained on the full dataset.  

These results further reinforce our assumption that continuous labels provide a improved strategy to balance datasets while accounting for both performance and fairness.

\begin{table*}[ht]
\renewcommand{\arraystretch}{1.0}
\setlength{\tabcolsep}{7pt}
\centering
\caption{ Results for iResNet-34 (ElasticArcFace) trained on several versions of the training set. Ranging from no sampling to three different sampling strategies, where two are based on Protocol C - \textbf{C} and \textbf{C(R)}. These latter strategies seem to hold significant advantages with respect to random sampling. Results shown for several variations of the number of identities kept in the training set. 
}
\label{tab:protocolc_elastic}
\begin{tabular}{lllrrrrrr}
\Xhline{2\arrayrulewidth}
      \#ID  & Strategy   &    \multicolumn{4}{c}{Accuracy per skin tone (\%) $\uparrow$} & Average &\multicolumn{2}{c}{Fairness $\downarrow$}\\
           \cmidrule(rl){3-6}\cmidrule(rl){8-9}  
   & & Caucasian      & \multicolumn{1}{c}{Indian} &\multicolumn{1}{c}{Asian} & \multicolumn{1}{c}{African} &  & \multicolumn{1}{c}{STD} & \multicolumn{1}{c}{SER}    \\
\hline
28k  & None &  96.67\%  & 94.88\% & 94.22\% & 93.38\%  & 94.79\%& 1.39 & 1.99     \\
\hline
\hline
27k & Random & 96.65\%  & 94.82\%  & 94.20\% & 93.77\% & \textbf{94.86}\% & \textbf{1.27} & 1.86 \\
\hline
24.5k & Random & 96.28\%  & 94.47\%  & 93.55\% & 93.15\% & 94.36\% & 1.39 & 1.84  \\
\hline
21k & Random & 95.78\%  & 93.97\%  & 93.20\% & 92.43\% & 93.85\% & \textbf{1.44}	 & \textbf{1.79} \\
\hline
14k & Random & 93.82\%  & 92.10\%  & 91.08\% & 90.00\% & 91.75\% & 1.62 & \textbf{1.62} \\
\hline
\hline
27k & C& 96.68\%  & 94.73\%  & 94.07\% & 93.93\% & 94.85\% & \textbf{1.27} & \textbf{1.83} \\
\hline
24.5k & C& 96.43\%  & 94.80\%  & 93.72\% & 93.53\% & 94.62\% & 1.33 & 1.81 \\
\hline
21k & C& 96.12\%  & 94.55\%  & 93.73\% & 92.28\% & 94.17\% & 1.60 & 1.99 \\
\hline
14k & C& 94.60\%  & 93.13\%  & 92.32\% & 90.85\% & \textbf{92.73}\% & \textbf{1.57} & 1.69  \\
\hline
\hline
27k & C(R)& 96.52\%  & 94.98\%  & 94.33\% & 93.32\% & 94.79\% & 1.34  & 1.92 \\
\hline
24.5k & C(R)& 96.18\%  & 94.78\%  & 94.22\% & 93.33\% & \textbf{94.63\%} & \textbf{1.19}  & \textbf{1.75} \\
\hline
21k & C(R)& 96.28\%  & 94.28\%  & 93.58\% & 92.87\% & \textbf{94.25\%} & 1.47  & 1.92 \\
\hline
14k & C(R)& 94.70\%  & 93.38\%  & 92.12\% & 90.73\% & \textbf{92.73\%} & 1.70 & 1.75 \\
\hline
\hline
\Xhline{2\arrayrulewidth}
\end{tabular}
\end{table*}

\subsubsection{Protocol C}

Following the previous analysis, we further explored the impact of the accounting for both the number of images and identities while balancing the dataset with our proposed score. In the other side of the spectrum, when compared to Protocol A, this Protocol C demonstrated a clear dominance of performance-related metrics when compared to bias-related ones. 

C achieves competitive results, performance-wise, with protocol B, and manages to surpass the model trained on the full dataset under some scenarios. However, it diminish the gap between random sampling and ethnicity oriented sampling with regards to bias metrics. This is seen in Table~\ref{tab:protocolc_elastic}, where, despite differences in overall performance of the models, bias metrics are similar between Random, and both \textbf{C} and \textbf{C(R)}. This results reinforce that even if we consider the number of identities and images per ethnicity group, the ethnicity score of each identity is far more relevant and meaningful.

\subsubsection{Comparison between Protocols}

To better visualize the trade-offs of considering different elements in the ethnicity-score based balancing of a dataset we propose the computation of a pareto frontier between all the trained models on each protocol. 

\begin{figure*}[t!]
    \centering
    % \begin{subfigure}[t]{0.325\linewidth}
    %     \centering
    %     \includegraphics[width=\linewidth]{plot-std.png}
    % \end{subfigure}%
    % ~
    \begin{subfigure}[t]{0.495\linewidth}
        \centering
        \includegraphics[width=\linewidth]{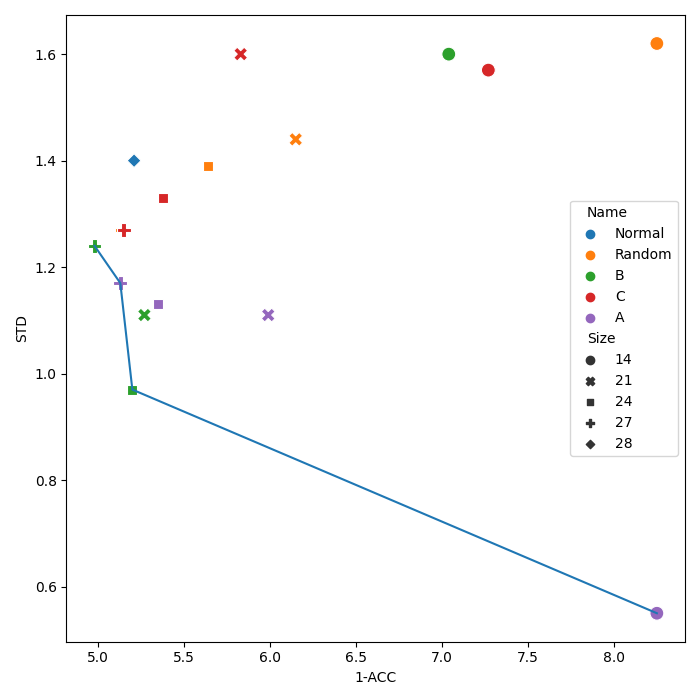}
        \caption{STD vs (1-ACC)}
        \label{fig:acc_std}
    \end{subfigure}
    \begin{subfigure}[t]{0.495\linewidth}
        \centering
        \includegraphics[width=\linewidth]{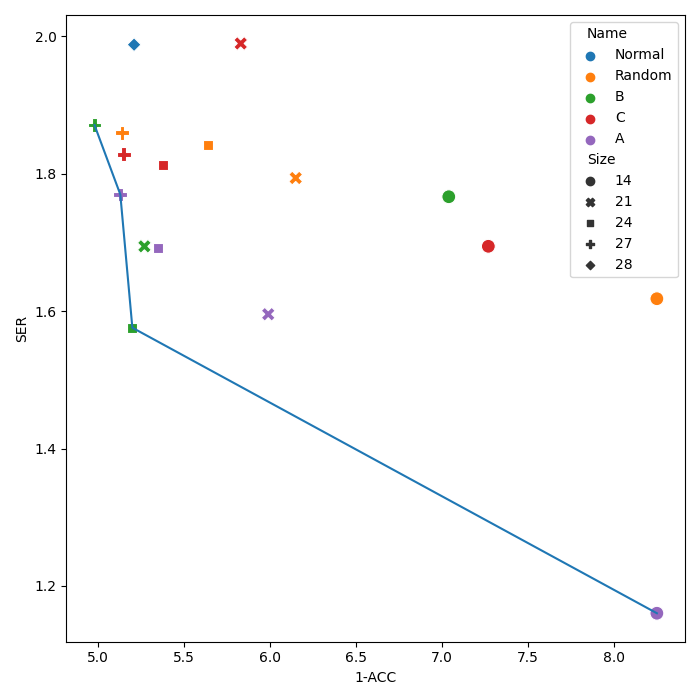}
        \caption{SER vs (1-ACC)}
        \label{fig:acc_seer}
    \end{subfigure}
    \caption{Trade-off between bias and performance visually highlighted through the computation of a pareto frontier with all the dominant solutions. Different shapes on the points represent different dataset sizes, while the colour reflects the sampling strategy. Pareto frontier elements are connected through a blue line. }
    \label{fig:pareto_optimality}
\end{figure*}

Figure~\ref{fig:pareto_optimality} shows the trade-off between the error (the complement of the accuracy) and STD, and also between the error and SER. On Figure~\ref{fig:acc_std} it is visible the superior performance of sampling strategies A and B, with the both having half of their models on the pareto frontier. These four models dominate all other instances, even the one without sampling. Random sampling shows inferior performance being the worse for all sampling magnitudes. Similar findings are displayed in Figure~\ref{fig:acc_seer}, where B and A outperform the remaining strategies. 

If one conducts a similar analysis per each dataset size, the results reflect the same conclusions with both A and B leading with regards to performance and bias. While B's performance is the best in all scenarios, A's dominance is only seen in three out of four dataset sizes.  

These results reinforce the findings of the previous section, and the relevance of the ethnicity score in the balancing of these datasets. As such, one can see that placing ethnicity labels in a continuous spectrum, while giving less relevance to the total number of image leads to fairer models that retain significantly more performance.

\section{Conclusion}
\label{sec:con}

During this work we have introduced the concept of an ethnicity flow, as ethnicity labels should be seen in a continuous space instead of the traditional discrete categorisation. We propose a strategy to calculate new labels for each of the images in the dataset, and we show that these are aligned with misclassified samples too. This continuous space reflected a clear presence of strong ethnicity features on higher scores, and less evident features on lower score samples.

Our proposed formulation is straightforward and can be validated by removing samples that unbalance the dataset. We showcased the superiority of this strategy when compared to discrete labels by comparing models trained on datasets balanced in the continuous space, and models trained on datasets that retain discrete balance. The performance and bias gap was significant in the majority of the scenarios with clear advantage for the continuous labels.  

We further release the set of samples produced by our sampling approach, and the code that can be used to replicate, reproduce and enhance the proposed methodology. In addition, algorithmic bias mitigation strategies can be applied on top of these set, hopefully boosting the performance and bias mitigation even further.

{\small
\bibliographystyle{ieee}
\bibliography{egbib}
}

\clearpage

\begin{table*}[h!]
\renewcommand{\arraystretch}{1.0}
\setlength{\tabcolsep}{7pt}
\centering
\caption{ Results for iResNet-50 (ElasticArcFace) trained on several versions of the training set. Ranging from no sampling to four different sampling strategies, A, B, C and Random. 
}
\label{chapter15_tab:res50}
\begin{tabular}{lllrrrrrr}
\Xhline{2\arrayrulewidth}
      \#ID  & Strategy   &    \multicolumn{4}{c}{Accuracy per skin tone (\%) $\uparrow$} & Average &\multicolumn{2}{c}{Fairness $\downarrow$}\\
           \cmidrule(rl){3-6}\cmidrule(rl){8-9}  
   & & Caucasian      & \multicolumn{1}{c}{Indian} &\multicolumn{1}{c}{Asian} & \multicolumn{1}{c}{African} &  & \multicolumn{1}{c}{STD} & \multicolumn{1}{c}{SER}    \\
\hline
28k  & None &  96.83\%  & 95.33\% & 95.00\% & 94.16\%  &  95.33\%&  1.11 & 1.84 \\
\hline
\hline
24.5k & Random & 96.32\%  & 94.73\% & 94.27\% & 93.67\%  & 94.75\%&   1.13	& 1.72 \\
\hline
24.5 & A & 96.35\%  & 95.02\% & 94.47\% & 93.88\%  & 94.93\%&  \textbf{1.05}	 & \textbf{1.68 }\\
\hline
24.5k & B& 97.02\%  & 95.00\%  & \textbf{94.93\%} & 94.05\% & \textbf{95.25\%} &  1.26 & 2.00 \\
\hline
24.5k & C& \textbf{96.87}\%  & \textbf{95.32\%}  & 94.33\% & \textbf{94.25\%} & 95.19\% &  1.22 &  1.84\\
\hline
\hline
\Xhline{2\arrayrulewidth}
\end{tabular}
\end{table*}

\begin{table*}[h!]
\renewcommand{\arraystretch}{1.0}
\setlength{\tabcolsep}{7pt}
\centering
\caption{ Results for iResNet-100 (ElasticArcFace) trained on several versions of the training set. Ranging from no sampling to four different sampling strategies, A, B, C and Random. 
}
\label{chapter15_tab:res100}
\begin{tabular}{lllrrrrrr}
\Xhline{2\arrayrulewidth}
      \#ID  & Strategy   &    \multicolumn{4}{c}{Accuracy per skin tone (\%) $\uparrow$} & Average &\multicolumn{2}{c}{Fairness $\downarrow$}\\
           \cmidrule(rl){3-6}\cmidrule(rl){8-9}  
   & & Caucasian      & \multicolumn{1}{c}{Indian} &\multicolumn{1}{c}{Asian} & \multicolumn{1}{c}{African} &  & \multicolumn{1}{c}{STD} & \multicolumn{1}{c}{SER}    \\
\hline
28k  & None &  97.12\%  & 95.78\% & 94.93\% & 95.36\%  &  95.80\% & 0.95 & 1.76\\
\hline
\hline
24.5k & Random & 97.08\%  & 95.72\% & 94.73\% & 94.87\%  & 95.60\%& 1.08  & 1.80\\
\hline
24.5 & A & 96.93\%  & 95.65\% & 95.13\% & 95.22\%  & 95.73\%& \textbf{0.83 } & \textbf{1.59} \\
\hline
24.5k & B& \textbf{97.27\%}  & 95.65\%  & \textbf{95.15\%} & 95.28\% & 95.84\% & 0.98	 & 1.78  \\
\hline
24.5k & C& 97.15\%  & \textbf{96.02\% } & 95.05\% & \textbf{95.38\%} & \textbf{95.90\%} & 0.93	 &  1.74\\
\hline
\hline
\Xhline{2\arrayrulewidth}
\end{tabular}
\end{table*}

\appendix
\section{Supplementary}

\subsection{Different Architectures}

In order to validate previous findings on larger architectures, we trained a ResNet-50, and a ResNet-100 on the data given by each protocol in addition to the random data. Due to the large number of potential experiments, we decided to not consider relabeling, and to only remove 3.5k samples. Nonetheless we firmly believe that previous experiments in addition to the ones presented here, constitute sufficient evidence of the effectiveness of continuous labels. 

ResNet-50 results are presented in Table~\ref{chapter15_tab:res50}. From these results, one can quickly infer a behavior similar to ResNet-34. When compared to our sampling approaches, random sampling is only close, yet with some difference, in performance to strategy A. However, this strategy achives better fairness metrics than Random sampling.  Strategies B and C seem to have lower fairness, but when carefully analysing them, their average accuracy is far higher than random sampling, and each individual demographic has improved performance in comparison.  

An analysis of the results of ResNet-100, shown in Table~\ref{chapter15_tab:res100}, leads to similar conclusions. However, in this case, strategies B and C not only surpass random in average accuracy, but achieve better values for the fairness metrics too. Moreover, they even surpass, with respect to average accuracy, the model trained on the full dataset, with C also achieving lower values of bias. These results further motivate the use of continuous labels to balance face recognition datasets.

\subsection{Different Loss}

In addition to validate on different architectures, we explored ResNet-34 with a different face recognition loss. This is motivated by the fact that, in this case, our Ethnicity classifier comes from a model pre-trained with ElasticArcFace. Hence, we repeat a large portion of the experiments with CurricularFace, highlighting that our sampled sets are effective, less biased and originate better models than sets balanced in the discrete space. 

\begin{table*}[h!]
\renewcommand{\arraystretch}{1.0}
\setlength{\tabcolsep}{7pt}
\centering
\caption{ Results for iResNet-34 (CurricularFace) trained on several versions of the training set. Ranging from no sampling to four different sampling strategies, A, B, C and Random. These latter strategies seem to hold significant advantages with respect to random sampling. Results shown for several variations of the number of identities kept in the training set. 
}
\label{chapter15_tab:curricular}
\begin{tabular}{lllrrrrrr}
\Xhline{2\arrayrulewidth}
      \#ID  & Strategy   &    \multicolumn{4}{c}{Accuracy per skin tone (\%) $\uparrow$} & Average &\multicolumn{2}{c}{Fairness $\downarrow$}\\
           \cmidrule(rl){3-6}\cmidrule(rl){8-9}  
   & & Caucasian      & \multicolumn{1}{c}{Indian} &\multicolumn{1}{c}{Asian} & \multicolumn{1}{c}{African} &  & \multicolumn{1}{c}{STD} & \multicolumn{1}{c}{SER}    \\
\hline
28k  & None & 96.55\%  & 94.90\% & 94.37\% & \textbf{93.97\%}  &  94.95\%& 1.13 & 1.75\\
\hline
\hline
27k & Random & 96.55\%  & 94.88\%  & 94.37\% & 93.75\% & 94.89\% & 1.20  & 1.81 \\
\hline
24.5k & Random & 96.27\%  & 94.72\% & 93.97\% & 93.30\%  & 94.57\%& 1.28  & 1.80\\
\hline
21k & Random & 95.50\%  & 94.15\%  & 93.28\% & 92.47\% & 93.85\% & 1.30  & 1.67 \\
\hline
14k & Random & 93.95\%  & 92.72\%  & 90.98\% & 89.87\% & 91.88\% & 1.81  & 1.67 \\
\hline
\hline
27k & A & \textbf{96.83\%}  & 94.93\% & 94.30\% & \textbf{94.03\%}  & \textbf{95.02\%}&  1.26 & 1.88\\
\hline
24.5 & A & 96.07\%  & 94.87\% & 94.10\% & 93.38\%  & 94.61\%&  \textbf{1.15} & \textbf{1.68} \\
\hline
21k & A & 95.22\%  & 94.10\% & 93.23\% & 93.05\%  & 93.90\%& \textbf{0.99} & \textbf{1.45}\\
\hline
14k & A & 92.17\%  & 92.33\% & 91.98\% & 91.25\%  & 91.93\% &  \textbf{0.48 }& \textbf{1.14}\\
\hline
\hline
27k & B& 96.43\%  & 95.15\%  & 94.25\% & 93.67\% & 94.88\% & 1.20 & \textbf{1.77 }\\
\hline
24.5k & B& \textbf{96.58\%}  & \textbf{94.95\%}  & \textbf{94.23\%} & \textbf{93.75\%} & \textbf{94.88\%} & 1.24  & 1.83 \\
\hline
21k & B& 96.10\%  & 94.47\%  & \textbf{94.15\%} & \textbf{93.10\%} & \textbf{94.46\%} &1.24  & 1.77 \\
\hline
14k & B& \textbf{95.10\%}  & \textbf{93.72\%}  & \textbf{92.25\%} & \textbf{91.72\%} & \textbf{93.20\%} & 1.52 & 1.69 \\
\hline
\hline
27k & C& 96.42\%  &\textbf{ 95.22\%}  & \textbf{94.37\%} & 93.68\% & 94.92\% & \textbf{1.18} & \textbf{1.77 }\\
\hline
24.5k & C& 96.22\%  & 94.85\%  & 94.07\% & 93.45\% & 94.65\% & 1.19  & 1.73 \\
\hline
21k & C& \textbf{96.20\%}  & \textbf{94.60\% } & 93.53\% & 92.67\% & 94.25\% & 1.52 & 1.93 \\
\hline
14k & C& 94.62\%  & \textbf{93.72\%}  & 92.07\% & 91.12\% & 92.88\% & 1.58 & 1.65 \\
\hline
\hline
\Xhline{2\arrayrulewidth}
\end{tabular}
\end{table*}

In Table~\ref{chapter15_tab:curricular}, it is visible that Random sampling was unable to surpass the remaining models on every dataset size and for all metrics, either performance or fairness ones. Furthermore, it is possible to note the large impact that strategy A has on the bias reduction, and that B has on the performance maintenance as the size of the training set reduces. More impressive is the fact that such performance maintenance does not make this model less fair than the ones trained on Random sampling. Having our hypothesis validated on a second face recognition loss further motivates its adoption in future research.

\end{document}